\documentclass{article}

\PassOptionsToPackage{numbers, compress}{natbib}

\usepackage[preprint]{neurips_2024}

\usepackage[utf8]{inputenc} 
\usepackage[T1]{fontenc}    
\usepackage{hyperref}       
\usepackage{url}            
\usepackage{booktabs}       
\usepackage{amsfonts, amssymb}       
\usepackage{amsmath}
\usepackage{nicefrac}       
\usepackage{microtype}      
\usepackage{xcolor}         
\usepackage{graphicx}
\title{Cascading Refinement Video Denoising with Uncertainty Adaptivity}

\author{%
  Xinyuan Yu \\
  \text{yxy1106@pku.edu.cn} \\
}

\begin{document}

\maketitle

\begin{abstract}
Accurate alignment is crucial for video denoising. However, estimating alignment in noisy environments is challenging. This paper introduces a cascading refinement video denoising method that can refine alignment and restore images simultaneously. Better alignment enables restoration of more detailed information in each frame. Furthermore, better image quality leads to better alignment. This method has achieved SOTA performance 
by a large margin on the CRVD \cite{yue2020supervised} dataset. Simultaneously, aiming to deal with multi-level noise, an uncertainty map was created after each iteration.
Because of this, redundant computation on the easily restored videos was avoided. By applying this method, the 
entire computation was reduced by 25\% on average.
\end{abstract}

\section{Introduction}
Serious noise inevitably appears in low-light frames, affecting video quality and influencing subsequent video analysis. Many tasks, such as visual odometry or tracking algorithms, are sensitive to the noise; however, recent video denoising algorithms still have many shortcoming.

A video denoising algorithm needs to estimate alignment and exploit redundant information from unaligned video frames for restoration; however, estimating accurate alignment is challenging even in clean images. Incorrect alignment will lead to artifacts in the restored frames. Moreover, when dealing with videos, alignment estimation errors easily propagate from frame to frame, increasing the adverse effects. Small or fast-moving objects, varying light conditions, and occlusion are all issues that seriously affect alignment performance. In addition, recent video denoising methods \cite{chan2021basicvsr, xue2019video} depend on techniques such as optical flow for alignment; these algorithms are sensitive to noise and, therefore, cannot achieve accurate alignments. A common solution is adding a pre-denoising module before alignment estimation; however, the pre-denoising module is usually a denoising method with a single frame or small
 window \cite{tassano2019fastdvdnet, yue2020supervised}. These methods are either unable to restore high-frequency details such as image texture, or they produce blurry edges. In our method, we introduce a cascading refinement structure, which can iteratively refine a restored image and alignment. 
 
After each iteration, the image quality is improved due to the removal of  noise and the addition of updated details, and the alignment is slightly refined. Better image quality leads to better alignment, and better alignment allows us to restore more detailed information in each frame.

Another challenge of video denoising is dealing with multi-level noise without prior noise information. The noise level depends on many factors such as ISO, lightness, temperature among others \cite{boncelet2009image}. It is usually unknown in practical applications. Even in the same image, the noise level happens to be different in different areas \cite{zhao2019enhancement}. Some of the existing deep learning-based denoising methods either require a single noise-level as input or depend on noise information as input. Some other methods \cite{yu2019deep}
add  extra network structures to handle multi-level noise, (e.g., \cite{yu2019deep} use several iterative down-up scaling 
modules). These modules become redundant for images with light noise. Our network adapts to multi-level noise by 
adding an uncertainty estimation module after each refinement block. We demonstrate that the uncertainty has a positive 
correlation with the restoration error in our experiment. Thus, the uncertainty can be used as a reference for whether to continue the refinement process. 

For the cascading refinement network, we adopt a recurrent iterative refinement structure for optical flow 
estimation. Following this, a flow-guided deformable convolution is used to provide diversity in offset estimation, thereby improving denoising performance. We fuse the refined feature map of the reference frame with the anligned feature maps of the supporting frames. The fused output serves as the input of the next iteration. After each iteration, both the uncertainty map and denoised patches are estimated. If the uncertainty is below a certain 
threshold, the output of the current iteration is directly set as the final output. Otherwise, we conduct another refinement iteration on the current frame.

This paper has the following contributions:

\begin{itemize}
  \item We design a cascading refinement method that can iteratively estimate both the alignment and restored image simultaneously. By using this structure, our method has achieved SOTA performance by a large margin on the CRVD dataset.

  \item Our network is adaptive to multi-level noise by using an uncertainty estimation module after each refinement block. This reduces the computation cost by 25\% on average.
\end{itemize} 

\section{Related Works}
\paragraph{Video Denoising}
Many early video denoising methods make assumptions regarding the noise distribution, such as assuming the noise follows a Gaussian distribution.\cite {buades2016patch,maggioni2011video,ji2010robust}; however, 
in practical application, the noise depends on many factors, making it difficult to meet this assumption. Toflow \cite{xue2019video} 
and DVDnet\cite{tassano2019dvdnet} are both methods based on optical flow. ViDeNN \cite{claus2019videnn} and DVDnet's fast
version, FastDVDnet\cite{tassano2019fastdvdnet}, use a two-stage spatial-temporal architecture. Optical flow is quite
sensitive to noise, making it difficult to achieve accurate alignment by using optical flow. 
 \cite{yue2020supervised} focuses on raw data and uses pyramid feature maps to estimate alignment. 
 \cite{chen2019seeing} published a Dark Raw Video (DRV) dataset. Their method is to deal with raw video 
 of dark scenes. However, none of the aforementioned works' networks are adaptive to different noise
 levels. Our network is unique in its design of cascading refinement and the uncertainty adaptivity. 
\paragraph{Iterative Refinement}
Our work is inspired by RAFT \cite{teed2020raft}, which introduces an iterative refinement structure for optical
flow. Prior to RAFT, several other methods, such as SpyNet \cite{ranjan2017optical}, PWC-Net \cite{sun2018pwc}, 
LiteFlowNet \cite{hui2018liteflownet} and VCN \cite{yang2019volumetric}, also applied iterative 
refinement; however, these works usually stack the network and are only able to iterate a few times.
Similar to RAFT, our work uses a lightweight update operator and shares weights between iterations. Therefore, 
we are able to run more iterations than other methods. Several other denoising methods \cite{yao2014iterative,luo2012generalized}
 also use iterative refinement methods; however, they are based on optimization.
It is difficult to design a robust hand-crafted objective. Some other methods \cite{adler2017solving, adler2018learned} 
learn the iterative process from data. In \cite{adler2017solving}, a gradient-like iterative scheme is introduced, where the gradient is learned using a convolutional network. Similar ideas are adopted in image denoising \cite{kobler2017variational},
tomographic reconstruction \cite{adler2018learned}, and novel view synthesis \cite{flynn2019deepview}. 
We are the first to use iterative refinement for the task of video denoising. 
\paragraph{Anytime Inference}
Network pruning \cite{han2015learning} and quantization \cite{rastegari2016xnor}, and distillation \cite{hinton2015distilling} are all 
techniques to reduce the computation cost. With these techniques, although the network computation is reduced, the enire network will still be computed.
Anytime inference \cite{dean1988analysis} accelarates the network by deciding how much to compute. The inference process stops if
the result is already accurate enough. Otherwise, the inference process continues until satisfactory result is predicted.
Recent any-time inference works are mainly in the field of classification \cite{huang2017multi} and semantic
 segmentation \cite{liu2016learning, liu2021anytime}. \cite{liu2021anytime} introduces an anytime inference framework
of semantic segmentation. The method decides whether to exit the framework based on the confidence of semantic segmentation. 
In our work, we estimate the uncertainty of our result. The inference process halts if the results is definite.

\section{The Proposed Method}

Our target is to restore a clean image from both its corresponding noisy frames (reference frames) and its unaligned neighboring frames (supporting frames).
Our method contains three parts:
\begin{itemize}
  \item Pre-denoising and Patch Matching
  \item Cascading Refinement
  \item Uncertainty Adaptivity
\end{itemize} 
We introduce a cascading refinement video denoising method that conducts an iterative estimation of both the alignment and restored image simultaneously. In addition, we predict an uncertainty map to estimate the error between the restored frame and the ground truth. By using the uncertainty map, we adjust the network in an adaptive way for multi-level noise.

\subsection{Pre-denoising and Patch Matching}

As mentioned in \cite{chen2018learning}, an alignment algorithm is sensitive to noise and is crucial for video denoising. By pre-denoising, we can have a more accurate first guess for the alignment, making the later iterative refinement easier. Following \cite{yue2020supervised}, we train a single-frame denoising model. The model has a U-Net \cite{ronneberger2015u} structure and is first pretrained on a synthetic dataset and then fine-tuned on the CRVD dataset. The synthetic dataset is generated by adding different levels of noise onto the MOTChallenge dataset\cite{milan2016mot16}. We select the same four videos from the MOTChallenge dataset as \cite{milan2016mot16}.

\par
Many classic methods for video denoising, such as EDVR \cite{wang2019edvr}, 
RViDeNet \cite{yue2020supervised}, and Tdan \cite{tian2020tdan}, are compute-intensive; however, an enlarged patch size will increase the computation cost dramatically. In the meantime, a small patch size will restrict the algorithm's ability to deal with fast-moving objects, since the same objects are likely to move out of the patch in nearby frames, making reasonable alignment impossible.

To deal with this problem, we first conduct a patch-matching process on the pre-denoised frames. The purpose of this process is to find the corresponding patch for each particular patch in the supporting frames. We adopt normalized cross-correlation for patch matching \cite{lewis1995fast}. The target function for patch matching is as follows:
\begin{equation}
  R(x,y)= \frac{\sum _{x',y'} (T(x',y') \cdot I(x+x',y+y'))}{\sqrt{\sum _{x',y'} T(x',y')^2 . \sum _{x',y'} I(x+x',y+y')^2}}
\end{equation}

This method is fast but may fail in cases of changing rotation or illumination; however, due to the small patch size, rotation and illumination changes are unlikely to happen. In our ablation study, we prove that patch matching is helpful for our algorithm.

\begin{figure}
  \begin{center}
    \includegraphics[width=0.6\textwidth]{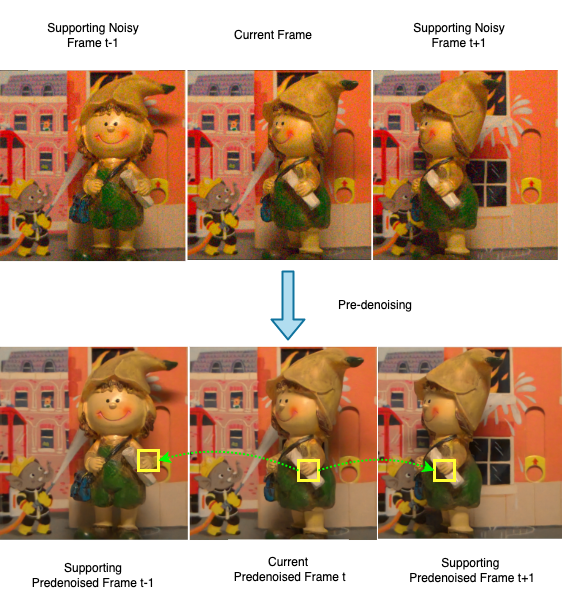}
    \caption{The process of patch matching is conducted on the pre-denoised images. The matched patches(indicated by yellow rectangle)
    are set as the input of the cascading refinement}
    \label{fig:pre_denoising}
  \end{center}
\end{figure}

The pre-denoising and patch matching process is illustrated in Fig.~\ref{fig:pre_denoising}. The patches are stacked
 and set as the input of the cascading refinement.

\subsection{Cascading Refinement}

\subsubsection{Iterative Refinement Optical Flow Estimation}

\begin{figure}
  \begin{center}
    \includegraphics[width=0.8\textwidth]{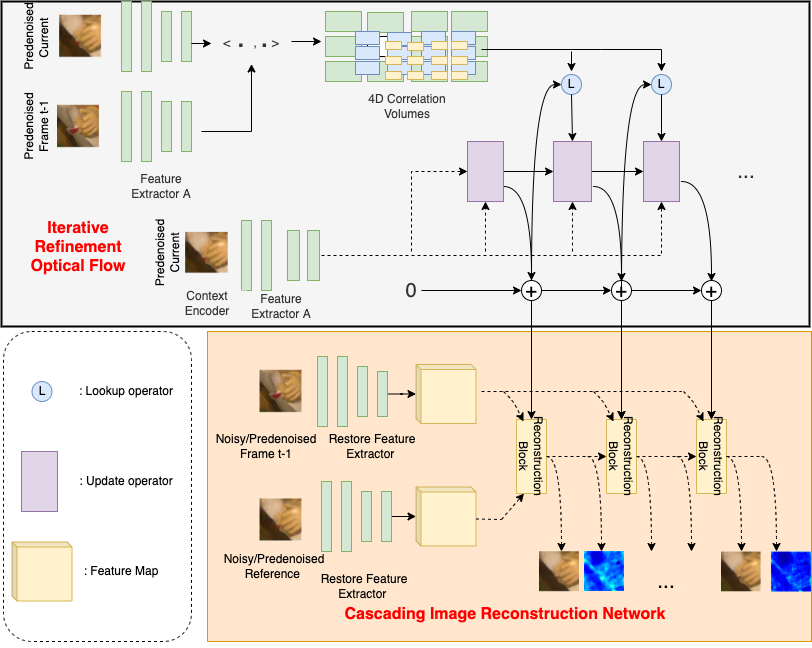}
    \caption{The iterative refinement optical flow estimation is shown as the upper part of this graph. The bottom part
    of this graph illustrates the cascading image reconstruction network. The network uses matched patches from three
     continuous frames as input. For simplicity, we only draw patches from two frames.}
    \label{fig:net_framework}
  \end{center}
\end{figure}

The structure of the cascading refinement module is shown in Fig. ~\ref{fig:net_framework}. Inspired by \cite{teed2020raft}, we use an iterative refinement method to estimate the optical flow between the pre-denoised 
patches. The optical flow estimation module is shown in the upper part of the graph. This module mainly contains three parts: (1). A feature extraction network, which extracts feature vectors from input patches. This network contains two parts: the feature encoder, which is used to extract feature vectors for computing the correlation between different frames, and the context encoder, which is used to extract features from the reference patch; (2) A 4D correlation volume, which represents the different scale correlation between each pair of pixels on the feature maps; (3) An update operator, which is used to iteratively refine the optical flow.

The features extracted by the context encoder of part (1) are injected directly into the update operator. Therefore, the output is able to better aggregate the spatial information from the reference patch. Each value of the
correlation volume is the inner product of the corresponding feature vectors as shown in Eq. \ref{similarity}, in which $g_{\theta}$ stands for the feature encoder, $g_{\theta}(I_{1}) \in \mathbb{R}^{H \times W \times D}$. $H, W, D$ 
stand for the height, width, and channel number, respectively. 

\begin{equation} 
  \label{similarity}
  C(g_{\theta}(I_{1}), g_{\theta}(I_{2})) \in \mathbb{R}^{H\times W\times H\times W}, \quad C_{ijkl}=\sum_{h}g_{\theta}(I_{1}).g_{\theta}(I_2)_{klh}
\end{equation}
We adopt a pyramid correlation volume as $\{C^1, C^2, C^3, C^4\}$, $C^1$ stands for the correlation volume with 
the original resolution of the feature map. ${C^2, C^3, C^4}$ represent the downsampled feature maps by average pooling. $C^k$ has the size of $ H \times W \times H/2^k \times W/2^k$. The pyramid correlation volume is used
to handle both large movement in the low resolution and small movement in the high resolution.

A correlation lookup operator is introduced in \cite{teed2020raft}. After estimating the optical flow, we update the index of each pixel in the warped feature map. The lookup operator is used to generate an updated feature map 
based on the updated index. The iterative refinement of the optical flow mimics the optimization process. Therefore, the weight of the update operator is shared between different iterations. The output of the update operator is a sequence of ${f_1, ..., f_N}$. The optical flow is initialized as $0$ and 
is updated after each iteration: $f_{k+1} = \bigtriangleup f+f_{k}$.
The update operator has a GRU \cite{cho2014properties} like structure. The fully connected layers are replaced by convolutions with a $3\times3$ kernel. The formula of the update operator is as follows:
\begin{equation} 
  z_{t} = \sigma  (Conv_{3\times3}([h_{t-1}, x_t], W_z))
\end{equation}
\begin{equation}
  r_{t} = \sigma  (Conv_{3\times3}([h_{t-1}, x_t], W_r))
\end{equation}
\begin{equation}
  \tilde{h_{t}} = tanh(Conv_{3\times3}([r_t\odot h_{t-1}], W_h))
\end{equation}
\begin{equation}
  h_{t} = (1-z_t) \odot h_{t-1} + z_t \odot \tilde{h_{t}} 
\end{equation}

$h$ stands for the hidden state, while $r$ and $z$ are the reset and update gating units. We add two convolutional layers after the hidden state $h$ to predict the flow update $\bigtriangleup f$.

\subsubsection{Cascading Image Reconstruction Network}

The inputs of the cascading image reconstruction network are stacked patches from the reference frame and the supporting frames. The pre-denoised patch and original noisy patch are both set as inputs since alignment estimation benefits from pre-denoised patches, and the original patches contain more high-frequency information, which is likely to be ignored by the pre-denoising process. A feature extractor is adopted for extracting features from the stacked original and pre-denoised patches separately for each frame. We use $f_\theta$ to represent the restoration feature extractor function. The restoration feature extractor is different from the feature extractor of the optical flow estimation network since the restoration process relies on different features from the optical flow estimation. The other input of the cascading image reconstruction network (CIRN) is the sequence of refined output from the optical flow estimation: $\{f_1,...,f_k\}$. For each iteration, we warp the feature maps of the supporting frames based on 
the optical flow of the current iteration. Then, we concatenate them with the reference frame's feature map. The feature map is refined iteratively by the reconstruction block.

Inspired by \cite{chan2021basicvsr++}, the reconstruction block contains a flow-guided deformable convolution(DCN) \cite{dai2017deformable} 
module after the input feature map. This module is to provide offset diversity. According to \cite{chan2020understanding}, the restoration performance is better with diverse offsets than with a single offset. We add a fusion block containing several residual blocks to fuse the aligned feature maps of supporting frames and the reference frame. Two heads are added after the fused feature maps for the purpose of image reconstruction and uncertainty estimation separately. Each head contains two convolutional layers. We will introduce the uncertainty estimation in section 3.2.3. The fused feature map is set as the input of the next reconstruction block. We use $m^{t}$ for the feature map of frame $t$. The optical flow $f_{k}^{i, j}$ stands for the optical flow estimated after iteration $k$ from frame $i$ to frame $j$. First, the feature map of the supporting frame is warped by the optical flow.
\begin{equation}
  \overline{m^{t-1}} = W(m^{t-1}, f_{k}^{t, t-1})
\end{equation}
$\overline{m^t}$ is the warped feature map of frame $t$. The input of flow-guided DCN is the optical flow, warped and original feature maps of the supporting frame, and feature map of the reference frame after iteration $k-1$. we calculated the aligned feature map as:
\begin{equation}
o_{k}^{t, t-1} = f_{k}^{t, t-1} + C^o(c(r_k, \overline{m_{t-1}})) 
\end{equation}
\begin{equation}
a_{k}^{t, t-1} = \sigma (C^m(c(r_k, \overline{m_{t-1}})))
\end{equation}
\begin{equation}
\hat{m_{t-1}} = D(m_{t-1}; o_{k}^{t, t-1}, a_{k}^{t, t-1})
\end{equation}
$C^{o,m}$ stands for several convolution layers, and $\sigma$ stands for the sigmoid function. $o_{k}^{t, t-1}$ and $a_{k}^{t, t-1}$
stand for DCN offset and mask. $r_k$ means the feature map of the reference patch after iteration $k$. $D$ denotes the deformable convolution. $c$ means the concat operation. $\hat{m_{t-1}}$ is the aligned feature map of frame $t-1$ after deformable convolution. We concatenate all the aligned feature maps of the supporting frames and $r_k$. In our case, we only use three frames.
\begin{equation} r_{k+1} = F(c(m_{t-1}, r_{k}, m_{t+1}))\end{equation}
$F$ denotes for the fusion block. We estimate the reconstructed patch and uncertainty map from $r_{k+1}$. $C^{r, u}$ stands for
several convolutional layers for reconstuction patch and uncertainty patch estimation.
\begin{equation}
s_{k+1} = C^{r}(r_{k+1}),
u_{k+1} = C^{u}(r_{k+1}) 
\end{equation}
$s_{k+1}$ and $u_{k+1}$ stand for the denoised patch and uncertainty map after the iteration $k+1$.

\begin{figure}
  \begin{center}
    \includegraphics[width=0.8\textwidth]{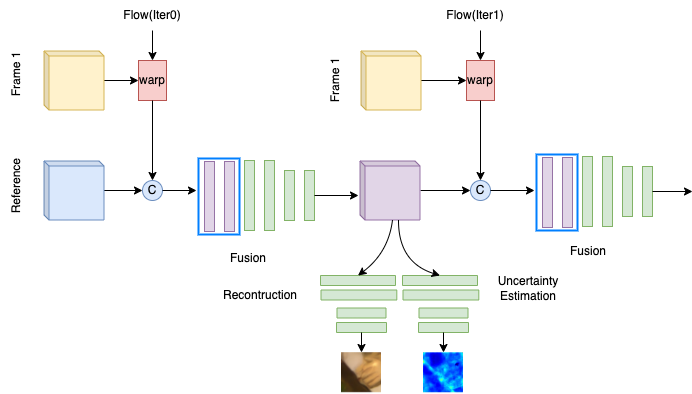}
    \caption{The reconstruction block contains a flow-guided deformable module. A fusion module is added to fuse the aligned feature maps of supporting frames and the reference frame. 
    Two heads follow the fused feature maps for image reconstruction and uncertainty estimation separately. The blue rectangle denotes deformable convolutions.}
    \label{fig:reconstruction_block}
  \end{center}
\end{figure}

\subsection{Uncertainty Adaptivity Module}
\begin{figure}
  \begin{center}
    \includegraphics[width=0.9\textwidth]{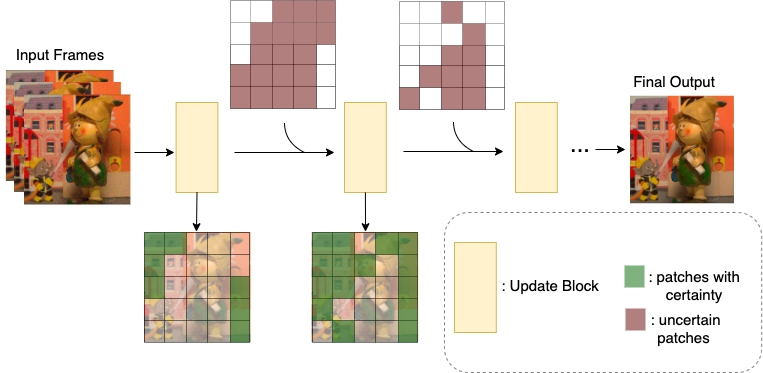}
    \caption{After each iteration, we calculate the mean value of the uncertainty map for each patch. If the mean value is larger than a certain threshold, we directly output the denoised result. Otherwise, we continue the iterative refinement procedure.}
    \label{fig:uncertainty_adaptivity}
  \end{center}
\end{figure}

Inspired by \cite{ning2021uncertainty}, we adopt an uncertainty-based loss. We assume the overall observation model as: 
\begin{equation}
  G_{i} = R_k(P_i) + \varepsilon \theta_{i}
\end{equation}
In the above formula, $P_i$ is the packed input patches, and $G_{i}$ is the ground truth of the denoised patch. $\theta_{i}$ is an additive term that stands for the aleatoric uncertainty. $R_k$ denotes the entire denoising
function for the $k$-th iteration. $\varepsilon$ denotes for the Laplace distribution with zero mean and unit variance. The maximum likelihood estimation can be formed as a minimization of the following loss function:
\begin{equation}
  L_{EU} = \frac{1}{N} \sum_{i=1}^{N}  \frac{ \left \|R_k(P_i) - G_i\right \|  _{2} }{2\sigma^{2}_i}+\frac{1}{2}ln\sigma_i^2 
\end{equation}
In our work, we estimate the $R_k(P_i)$ and $\sigma^{2}$ simultaneously. We use $\sigma^2$ to judge whether to continue the iterative refinement.
The validity of the uncertainty estimation is proven in the experiment section. The entire process is illustrated in ~\ref{fig:uncertainty_adaptivity}

After each iteration, we calculate the mean value of the uncertainty map. If the mean value exceeds a certain threshold, we directly output the denoised result. Otherwise, we continue the iterative refinement procedure. This approach allows our network to adapt to inputs with varying levels of noise. In our experiment, we demonstrate that this method can save 25\% of computational cost for the CRVD dataset.
 \paragraph{Overall Loss} 
 We calculate the loss for the entire sequence of predictions with exponentially increasing weights. $L_{EU}^k$ stands for the $L_{EU}$ loss for  the $k$th iteration. $N$ denotes the maximum possible iterations, which we set to 12 in our experiment. The total loss is defined as follows:
 \begin{equation}
 	L = \sum_{k=1}^{N}\gamma^{N-k}L_{EU}^k  .
 \end{equation}

 \begin{table*}[ht]
   \begin{center}
     \begin{tabular}{cccccccc}
       \hline
       {Method} & {Noisy} & {VBM4D} & {TOFlow} & {Tdan} & {BasicVSR++} & {EDVR} & {Ours} \\
       \toprule
       {PSNR}  & 31.79 & 34.81 & 37.00 & 38.46 & 38.66 & 38.97  & \textbf{39.62} \\
       \hline
       {SSIM} & 0.752 & 0.922 & 0.962 & 0.972 & 0.970 & 0.972  & \textbf{0.978} \\
       \hline
     \end{tabular}
     \caption{Comparison of state-of-the-art methods on the CRVD sRGB dataset. The best results are highlighted in bold.}
     \label{tab:metrics}
   \end{center}
 \end{table*}
 
 \begin{figure*}
   \begin{center}
     \includegraphics[width=0.9\textwidth]{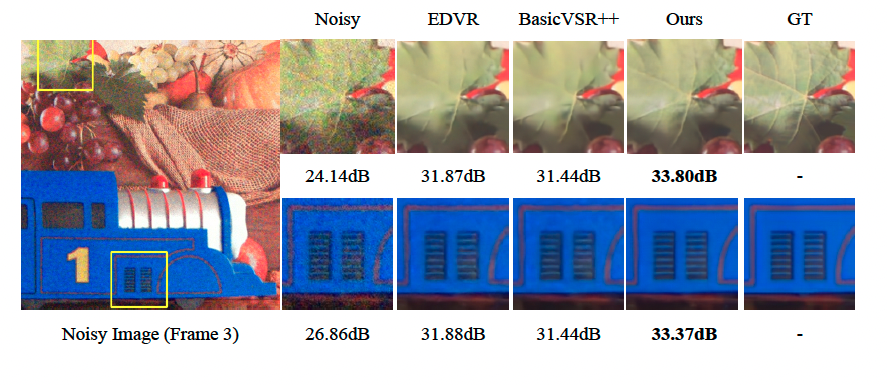}
     \caption{Comparison on a CRVD indoor scene with ISO 25600. Zoom in for better observation.}
     \label{fig:visual_plot1}
   \end{center}
 \end{figure*}
 
 \section{Experiments}
 We evaluate our method on the CRVD dataset \cite{yue2020supervised}, a realistic noisy-clean video dataset. It provides videos under different ISO settings. Our method achieves state-of-the-art performance on this dataset.
 
 \paragraph{Training Details}
 
 The input of our method is sRGB frames. We use only three frames as input (the current frame, the frame before, and the frame after). The input patch size of our work is $128\times128$, maintaining consistency with \cite{yue2020supervised}. \cite{yue2020supervised} uses a patch size of $256\times256$ on raw data, which is the same as ours after converting to sRGB.
 
 The iterative refinement optical flow estimation module is first pre-trained on FlyingChair \cite{dosovitskiy2015flownet} 
 and FlyingThings3D \cite{mayer2016large} for 100k iterations, then fine-tuned on Sintel \cite{butler2012naturalistic} 
 and KITTI-2015 \cite{geiger2013vision} for another 100k iterations, similar to \cite{teed2020raft}. 
 The pre-denoised model is also pre-trained on both a synthetic dataset and a realistic dataset similar to \cite{yue2020supervised}. 
 The parameters of these two pre-trained models are frozen for the first 30,000 iterations. 
 The other parts of our network are initialized with random weights. 
 The update operators share weights between iterations. 
 We train our model for a total of 700,000 iterations using the AdamW optimizer with a learning rate of 0.00002.
 We train our model on an Nvidia GeForce 3090 GPU with a batch size of 1. 
 
 We set the maximum iteration to 12. The threshold for stopping the refinement process is 0.002.

 \begin{figure*}
   \begin{center}
     \includegraphics[width=0.9\textwidth]{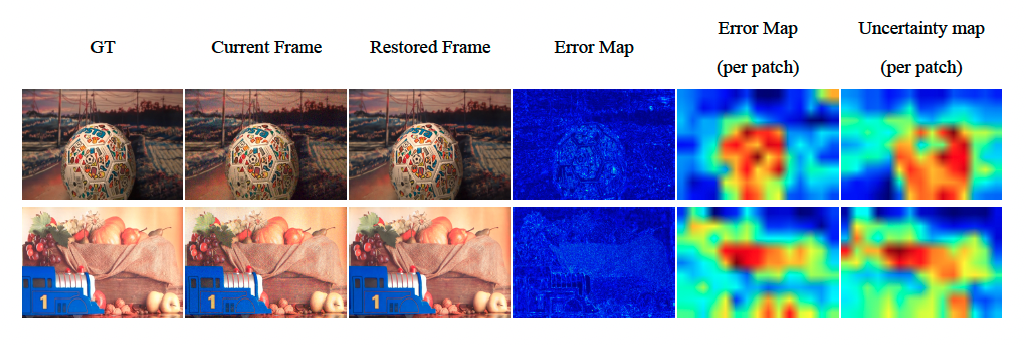}
     \caption{This graph presents the visualized results of scene 7 and scene 10 of the CRVD dataset. We show the error map, which is the absolute error between the ground truth and the restored frame. The two heat maps represent the upsampled average mean error and uncertainty per patch. }
     \label{fig:heat_map}
   \end{center}
 \end{figure*}

 \begin{figure}
   \begin{center}
      \includegraphics[width=0.4\textwidth, height=5cm]{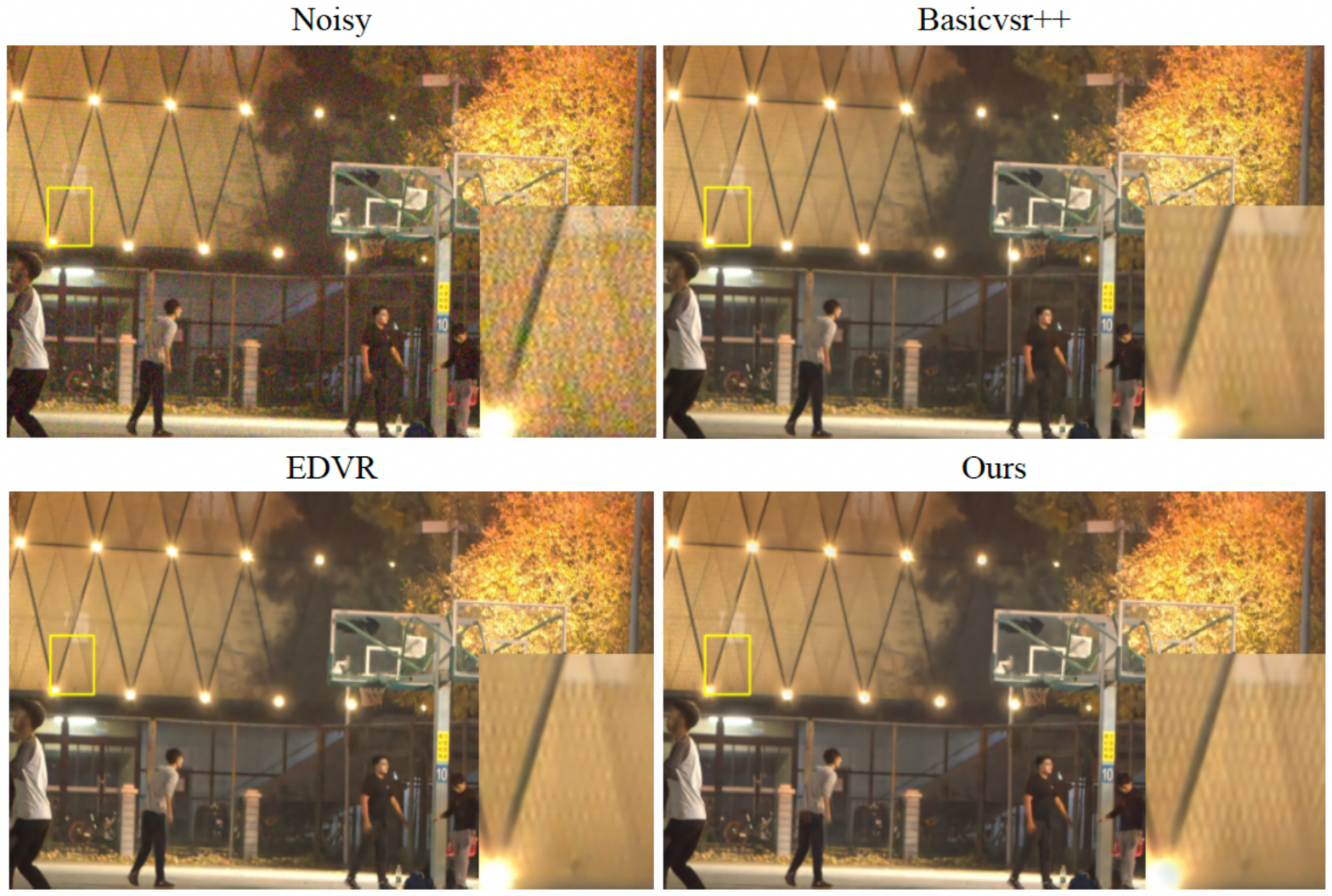}
      \caption{Visual quality comparison for frames of outdoor scenes. Zoom in for better observation.}
      \label{fig:outdoor}
   \end{center}
\end{figure}

 \subsection{Comparison with State-of-the-art Methods}
 
 We compare our method with several state-of-the-art video denoising methods. We test our method on the CRVD sRGB dataset, using the average of PSNR and SSIM for all 25 sequences of CRVD. All methods are trained on noisy-clean pairs. Our method improved the PSNR by 0.65 dB, as illustrated in Table~\ref{tab:metrics}.

 A visual quality comparison is given in Figure~\ref{fig:visual_plot1}. In the first row, we can see that our method retains more veins than other methods. Our method is more effective in retaining edges due to more accurate alignment. In the second row, color spots appear in smooth areas in the results of other methods. Our method has removed the noise more thoroughly due to the iterative refinement. We remove the noise progressively.

 To demonstrate the generalization ability of the model, we test it on the CRVD dataset outdoor scenes with high ISO values. 
 However, we don't have the ground truth for the outdoor scenes, so we compare the visual quality with other methods. 
 Figure~\ref{fig:outdoor} presents the performance on outdoor scenes. We can see that our method restores more textures compared to other methods.

 \subsection{Uncertainty Adaptivity}
 To compare the uncertainty map and error map. We provide the visualized heat map of average error and uncertainty per patch, as shown in Figure~\ref{fig:heat_map}.

 In the first row, we can see the error map and uncertainty map are both relatively large in the texture-rich area. In the second row, the error is large at the boundary and in the fabric with rich texture. The uncertainty map is also relatively high there. The uncertainty map and error map are spatially consistent.
 
\begin{figure}[t]
	\begin{center}
		\includegraphics[width=0.4\textwidth, height=4cm]{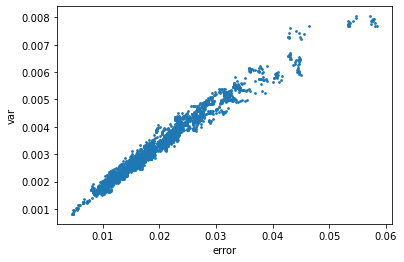}
		\caption{This graph is an illustration of the correlation between error and uncertainty}
		\label{fig:uncertainty}
	\end{center}
\end{figure}

 To be more precise, we calculate the Pearson correlation coefficient between the mean absolute error and mean uncertainty of each patch. Pearson's r is 0.984. This indicates a strong correlation between the error and uncertainty. The scatter plot of error and uncertainty is shown in Figure~\ref{fig:uncertainty}

 \subsection{Ablation Study}
 In this part, we conduct experiments to prove the effectiveness of cascading refinement, template matching 
 , and uncertainty adaptivity. We compare the PSNR and SSIM of the first iteration and the maximum iteration (12 iterations in our setting).
 The results show that the iterative refinement can provide an improvement in PSNR of about 0.4 dB. The result is slightly lower without the patch matching module. The patch matching module gives the method the ability to handle large motion. However, it also has the possibility to fail. Thus, patch matching improves performance but not significantly.
 To prove the effectiveness of uncertainty adaptivity, we conduct an experiment comparing the fixed iteration strategy (where the number of iterations is hard to set to 1)
 with a method using uncertainty adaptivity. With uncertainty adaptivity, the performance is almost the same; however, the average number of iterations is 75\% of the previous number. 
 Thus, we can save 25\% of the computation cost in the refinement.  
 
 \begin{table}[ht]
   \begin{center}
     \begin{tabular}{ccc} 
       \hline
        & {PSNR} & {SSIM} \\
       \toprule
       {Ours}  & 39.62 & 0.978 \\
       \hline
       {w/o cascading refinement} & 39.22 & 0.974 \\
       \hline
       {w/o template matching} & 39.51 & 0.977 \\
       \hline
       {With uncertainty adaptivity*} & 39.58 & 0.977 \\
       \hline
     \end{tabular}
     \caption{
       In other experiments, we use 12 iterations. In the experiment with uncertainty adaptivity, the average iteration number is 9.4. The performance is still comparable with fully-iterative experiments.
     }
   \end{center}
   \label{tab:ablation}
 \end{table}

 \section{Conclusion}
 
We have proposed a cascading refinement video denoising method that can refine alignment and restore images simultaneously.  We also estimate an uncertainty map after each iteration. Because of this, redundant computations on the effectively restored videos are avoided. Compared with other video denoising methods, our novelty lies in the design of the 
 cascading refinement and the uncertainty adaptivity. In our experiment, we showed that our algorithm outperforms existing video denoising methods.  With the uncertainty, our network is adaptive and flexible for multi-level noise.

{\small
\bibliographystyle{ieee}
\bibliography{egbib}

\begin{thebibliography}{10}\itemsep=-1pt

\bibitem{adler2017solving}
J.~Adler and O.~{\"O}ktem.
\newblock Solving ill-posed inverse problems using iterative deep neural networks.
\newblock {\em Inverse Problems}, 33(12):124007, 2017.

\bibitem{adler2018learned}
J.~Adler and O.~{\"O}ktem.
\newblock Learned primal-dual reconstruction.
\newblock {\em IEEE transactions on medical imaging}, 37(6):1322--1332, 2018.

\bibitem{boncelet2009image}
C.~Boncelet.
\newblock Image noise models.
\newblock In {\em The essential guide to image processing}, pages 143--167. Elsevier, 2009.

\bibitem{buades2016patch}
A.~Buades, J.-L. Lisani, and M.~Miladinovi{\'c}.
\newblock Patch-based video denoising with optical flow estimation.
\newblock {\em IEEE Transactions on Image Processing}, 25(6):2573--2586, 2016.

\bibitem{butler2012naturalistic}
D.~J. Butler, J.~Wulff, G.~B. Stanley, and M.~J. Black.
\newblock A naturalistic open source movie for optical flow evaluation.
\newblock In {\em European conference on computer vision}, pages 611--625. Springer, 2012.

\bibitem{chan2020understanding}
K.~C. Chan, X.~Wang, K.~Yu, C.~Dong, and C.~C. Loy.
\newblock Understanding deformable alignment in video super-resolution.
\newblock {\em arXiv preprint arXiv:2009.07265}, 4(3):4, 2020.

\bibitem{chan2021basicvsr}
K.~C. Chan, X.~Wang, K.~Yu, C.~Dong, and C.~C. Loy.
\newblock Basicvsr: The search for essential components in video super-resolution and beyond.
\newblock In {\em Proceedings of the IEEE/CVF Conference on Computer Vision and Pattern Recognition}, pages 4947--4956, 2021.

\bibitem{chan2021basicvsr++}
K.~C. Chan, S.~Zhou, X.~Xu, and C.~C. Loy.
\newblock Basicvsr++: Improving video super-resolution with enhanced propagation and alignment.
\newblock {\em arXiv preprint arXiv:2104.13371}, 2021.

\bibitem{chen2019seeing}
C.~Chen, Q.~Chen, M.~N. Do, and V.~Koltun.
\newblock Seeing motion in the dark.
\newblock In {\em Proceedings of the IEEE/CVF International Conference on Computer Vision}, pages 3185--3194, 2019.

\bibitem{chen2018learning}
C.~Chen, Q.~Chen, J.~Xu, and V.~Koltun.
\newblock Learning to see in the dark.
\newblock In {\em Proceedings of the IEEE conference on computer vision and pattern recognition}, pages 3291--3300, 2018.

\bibitem{cho2014properties}
K.~Cho, B.~Van~Merri{\"e}nboer, D.~Bahdanau, and Y.~Bengio.
\newblock On the properties of neural machine translation: Encoder-decoder approaches.
\newblock {\em arXiv preprint arXiv:1409.1259}, 2014.

\bibitem{claus2019videnn}
M.~Claus and J.~van Gemert.
\newblock Videnn: Deep blind video denoising.
\newblock In {\em Proceedings of the IEEE/CVF Conference on Computer Vision and Pattern Recognition Workshops}, pages 0--0, 2019.

\bibitem{dai2017deformable}
J.~Dai, H.~Qi, Y.~Xiong, Y.~Li, G.~Zhang, H.~Hu, and Y.~Wei.
\newblock Deformable convolutional networks.
\newblock In {\em Proceedings of the IEEE international conference on computer vision}, pages 764--773, 2017.

\bibitem{dean1988analysis}
T.~L. Dean and M.~S. Boddy.
\newblock An analysis of time-dependent planning.
\newblock In {\em AAAI}, volume~88, pages 49--54, 1988.

\bibitem{dosovitskiy2015flownet}
A.~Dosovitskiy, P.~Fischer, E.~Ilg, P.~Hausser, C.~Hazirbas, V.~Golkov, P.~Van Der~Smagt, D.~Cremers, and T.~Brox.
\newblock Flownet: Learning optical flow with convolutional networks.
\newblock In {\em Proceedings of the IEEE international conference on computer vision}, pages 2758--2766, 2015.

\bibitem{flynn2019deepview}
J.~Flynn, M.~Broxton, P.~Debevec, M.~Du-Vall, G.~Fyffe, R.~S. Overbeck, N.~Snavely, and R.~Tucker.
\newblock Deepview: High-quality view synthesis by learned gradient descent.
\newblock In {\em Conference on Computer Vision and Pattern Recognition (CVPR)}, volume~3, 2019.

\bibitem{geiger2013vision}
A.~Geiger, P.~Lenz, C.~Stiller, and R.~Urtasun.
\newblock Vision meets robotics: The kitti dataset.
\newblock {\em The International Journal of Robotics Research}, 32(11):1231--1237, 2013.

\bibitem{han2015learning}
S.~Han, J.~Pool, J.~Tran, and W.~Dally.
\newblock Learning both weights and connections for efficient neural network.
\newblock {\em Advances in neural information processing systems}, 28, 2015.

\bibitem{hinton2015distilling}
G.~Hinton, O.~Vinyals, J.~Dean, et~al.
\newblock Distilling the knowledge in a neural network.
\newblock {\em arXiv preprint arXiv:1503.02531}, 2(7), 2015.

\bibitem{huang2017multi}
G.~Huang, D.~Chen, T.~Li, F.~Wu, L.~Van Der~Maaten, and K.~Q. Weinberger.
\newblock Multi-scale dense networks for resource efficient image classification.
\newblock {\em arXiv preprint arXiv:1703.09844}, 2017.

\bibitem{hui2018liteflownet}
T.-W. Hui, X.~Tang, and C.~C. Loy.
\newblock Liteflownet: A lightweight convolutional neural network for optical flow estimation.
\newblock In {\em Proceedings of the IEEE conference on computer vision and pattern recognition}, pages 8981--8989, 2018.

\bibitem{ji2010robust}
H.~Ji, C.~Liu, Z.~Shen, and Y.~Xu.
\newblock Robust video denoising using low rank matrix completion.
\newblock In {\em 2010 IEEE Computer Society Conference on Computer Vision and Pattern Recognition}, pages 1791--1798. IEEE, 2010.

\bibitem{kobler2017variational}
E.~Kobler, T.~Klatzer, K.~Hammernik, and T.~Pock.
\newblock Variational networks: connecting variational methods and deep learning.
\newblock In {\em German conference on pattern recognition}, pages 281--293. Springer, 2017.

\bibitem{lewis1995fast}
J.~Lewis.
\newblock Fast normalized cross-correlation.
\newblock {\em Vision Interface}, 10, 1995.

\bibitem{liu2016learning}
B.~Liu and X.~He.
\newblock Learning dynamic hierarchical models for anytime scene labeling.
\newblock In {\em European Conference on Computer Vision}, pages 650--666. Springer, 2016.

\bibitem{liu2021anytime}
Z.~Liu, Z.~Xu, H.-J. Wang, T.~Darrell, and E.~Shelhamer.
\newblock Anytime dense prediction with confidence adaptivity.
\newblock In {\em International Conference on Learning Representations}, 2021.

\bibitem{luo2012generalized}
E.~Luo, S.~Pan, and T.~Nguyen.
\newblock Generalized non-local means for iterative denoising.
\newblock In {\em 2012 Proceedings of the 20th European Signal Processing Conference (EUSIPCO)}, pages 260--264. IEEE, 2012.

\bibitem{maggioni2011video}
M.~Maggioni, G.~Boracchi, A.~Foi, and K.~Egiazarian.
\newblock Video denoising using separable 4d nonlocal spatiotemporal transforms.
\newblock In {\em Image Processing: Algorithms and Systems IX}, volume 7870, page 787003. International Society for Optics and Photonics, 2011.

\bibitem{mayer2016large}
N.~Mayer, E.~Ilg, P.~Hausser, P.~Fischer, D.~Cremers, A.~Dosovitskiy, and T.~Brox.
\newblock A large dataset to train convolutional networks for disparity, optical flow, and scene flow estimation.
\newblock In {\em Proceedings of the IEEE conference on computer vision and pattern recognition}, pages 4040--4048, 2016.

\bibitem{milan2016mot16}
A.~Milan, L.~Leal-Taix{\'e}, I.~Reid, S.~Roth, and K.~Schindler.
\newblock Mot16: A benchmark for multi-object tracking.
\newblock {\em arXiv preprint arXiv:1603.00831}, 2016.

\bibitem{ning2021uncertainty}
Q.~Ning, W.~Dong, X.~Li, J.~Wu, and G.~Shi.
\newblock Uncertainty-driven loss for single image super-resolution.
\newblock {\em Advances in Neural Information Processing Systems}, 34, 2021.

\bibitem{ranjan2017optical}
A.~Ranjan and M.~J. Black.
\newblock Optical flow estimation using a spatial pyramid network.
\newblock In {\em Proceedings of the IEEE conference on computer vision and pattern recognition}, pages 4161--4170, 2017.

\bibitem{rastegari2016xnor}
M.~Rastegari, V.~Ordonez, J.~Redmon, and A.~Farhadi.
\newblock Xnor-net: Imagenet classification using binary convolutional neural networks.
\newblock In {\em European conference on computer vision}, pages 525--542. Springer, 2016.

\bibitem{ronneberger2015u}
O.~Ronneberger, P.~Fischer, and T.~Brox.
\newblock U-net: Convolutional networks for biomedical image segmentation.
\newblock In {\em International Conference on Medical image computing and computer-assisted intervention}, pages 234--241. Springer, 2015.

\bibitem{sun2018pwc}
D.~Sun, X.~Yang, M.-Y. Liu, and J.~Kautz.
\newblock Pwc-net: Cnns for optical flow using pyramid, warping, and cost volume.
\newblock In {\em Proceedings of the IEEE conference on computer vision and pattern recognition}, pages 8934--8943, 2018.

\bibitem{tassano2019dvdnet}
M.~Tassano, J.~Delon, and T.~Veit.
\newblock Dvdnet: A fast network for deep video denoising.
\newblock In {\em 2019 IEEE International Conference on Image Processing (ICIP)}, pages 1805--1809. IEEE, 2019.

\bibitem{tassano2019fastdvdnet}
M.~Tassano, J.~Delon, and T.~Veit.
\newblock Fastdvdnet: Towards real-time video denoising without explicit motion estimation.
\newblock In {\em IEEE/CVF Conference on Computer Vision and Pattern Recognition (CVPR)}, 2020.

\bibitem{teed2020raft}
Z.~Teed and J.~Deng.
\newblock Raft: Recurrent all-pairs field transforms for optical flow.
\newblock In {\em European conference on computer vision}, pages 402--419. Springer, 2020.

\bibitem{tian2020tdan}
Y.~Tian, Y.~Zhang, Y.~Fu, and C.~Xu.
\newblock Tdan: Temporally-deformable alignment network for video super-resolution.
\newblock In {\em Proceedings of the IEEE/CVF Conference on Computer Vision and Pattern Recognition}, pages 3360--3369, 2020.

\bibitem{wang2019edvr}
X.~Wang, K.~C. Chan, K.~Yu, C.~Dong, and C.~Change~Loy.
\newblock Edvr: Video restoration with enhanced deformable convolutional networks.
\newblock In {\em Proceedings of the IEEE/CVF Conference on Computer Vision and Pattern Recognition Workshops}, pages 0--0, 2019.

\bibitem{xue2019video}
T.~Xue, B.~Chen, J.~Wu, D.~Wei, and W.~T. Freeman.
\newblock Video enhancement with task-oriented flow.
\newblock {\em International Journal of Computer Vision}, 127(8):1106--1125, 2019.

\bibitem{yang2019volumetric}
G.~Yang and D.~Ramanan.
\newblock Volumetric correspondence networks for optical flow.
\newblock {\em Advances in neural information processing systems}, 32, 2019.

\bibitem{yao2014iterative}
X.-R. Yao, W.-K. Yu, X.-F. Liu, L.-Z. Li, M.-F. Li, L.-A. Wu, and G.-J. Zhai.
\newblock Iterative denoising of ghost imaging.
\newblock {\em Optics Express}, 22(20):24268--24275, 2014.

\bibitem{yu2019deep}
S.~Yu, B.~Park, and J.~Jeong.
\newblock Deep iterative down-up cnn for image denoising.
\newblock In {\em Proceedings of the IEEE/CVF Conference on Computer Vision and Pattern Recognition Workshops}, pages 0--0, 2019.

\bibitem{yue2020supervised}
H.~Yue, C.~Cao, L.~Liao, R.~Chu, and J.~Yang.
\newblock Supervised raw video denoising with a benchmark dataset on dynamic scenes.
\newblock In {\em Proceedings of the IEEE/CVF Conference on Computer Vision and Pattern Recognition}, pages 2301--2310, 2020.

\bibitem{zhao2019enhancement}
R.~Zhao, K.-M. Lam, and D.~P. Lun.
\newblock Enhancement of a cnn-based denoiser based on spatial and spectral analysis.
\newblock In {\em 2019 IEEE International Conference on Image Processing (ICIP)}, pages 1124--1128. IEEE, 2019.

\end{thebibliography}
}

\end{document}